\documentclass[11pt,a4paper]{article}
\usepackage[hyperref]{emnlp-ijcnlp-2019}
\usepackage{times}
\usepackage{latexsym}
\usepackage{graphicx}  
\usepackage[utf8]{inputenc}
\usepackage[french]{babel}

\usepackage{url}

\aclfinalcopy 

\title{Learning to Create Sentence Semantic Relation Graphs for Multi-Document Summarization}

\author{Diego Antognini \and
  Boi Faltings\\
  Artificial Intelligence Laboratory \\
  École Polytechnique Fédérale de Lausanne \\
    Lausanne, Switzerland \\
  {\tt \{diego.antognini, boi.faltings\}@epfl.ch} \\}

\date{}

\begin{document}
\maketitle
\begin{abstract}
Linking facts across documents is a challenging task, as the language used to express the same information in a sentence can vary significantly, which complicates the task of multi-document summarization.
Consequently, existing approaches heavily rely on hand-crafted features, which are domain-dependent and hard to craft, or additional annotated data, which is costly to gather.
To overcome these limitations, we present a novel method, which makes use of two types of sentence embeddings: universal embeddings, which are trained on a large unrelated corpus, and domain-specific embeddings, which are learned during training.
 To this end, we develop \textit{SemSentSum}, a fully data-driven model able to leverage both types of sentence embeddings by building a sentence semantic relation graph.
\textit{SemSentSum} achieves competitive results on two types of summary, consisting of 665 bytes and 100 words.
Unlike other state-of-the-art models, neither hand-crafted features nor additional annotated data are necessary, and the method is easily adaptable for other tasks. 
To our knowledge, we are the first to use multiple sentence embeddings for the task of multi-document summarization.
\end{abstract}

\section{Introduction}

Today's increasing flood of information on the web creates a need for automated multi-document summarization systems that produce high quality summaries.
However, producing summaries in a multi-document setting is difficult, as the language used to display the same information in a sentence can vary significantly, making it difficult for summarization models to capture.
Given the complexity of the task and the lack of datasets, most researchers use extractive summarization, where the final summary is composed of existing sentences in the input documents. 
More specifically, extractive summarization systems output summaries in two steps: via sentence ranking, where an importance score is assigned to each sentence, and via the subsequent sentence selection, where the most appropriate sentence is chosen, by considering 1)~their importance and 2)~their frequency among all documents.
Due to data sparcity, models heavily rely on well-designed features at the word level \cite{hong2014improving,cao2015ranking,christensen2013towards,Yasunaga17} or take advantage of other large, manually annotated datasets and then apply transfer learning \cite{CaoLLW17}. 
Additionally, most of the time, all sentences in the same collection of documents are processed independently and therefore, their relationships are lost. 

In realistic scenarios, features are hard to craft, gathering additional annotated data is costly, and the large variety in expressing the same fact cannot be handled by the use of word-based features only, as is often the case.
In this paper, we address these obstacles by proposing to simultaneously leverage two types of sentence embeddings, namely embeddings pre-trained on a large corpus that capture a variety of meanings and domain-specific embeddings learned during training. 
The former is typically trained on an unrelated corpus composed of high quality texts, allowing to cover additional contexts for each encountered word and sentence. 
Hereby, we build on the assumption that sentence embeddings capture both the syntactic and semantic content of sentences. 
We hypothesize that using two types of sentence embeddings, general and domain-specific, is beneficial for the task of multi-document summarization, as the former captures the most common semantic structures from a large, general corpus, while the latter captures the aspects related to the domain.

We present \textit{SemSentSum} (Figure~\ref{fig:architecture}), a fully data-driven summarization system, which does not depend on hand-crafted features, nor additional data, and is thus domain-independent. 
It first makes use of general sentence embedding knowledge to build a sentenc semantic relation graph that captures sentence similarities (Section~\ref{ssrg}). 
In a second step, it trains genre-specific sentence embeddings related to the domains of the collection of documents, by utilizing a sentence encoder (Section~\ref{sub:sentence_encoder}). 
Both representations are afterwards merged, by using a graph convolutional network \cite{KipfW16} (Section~\ref{sub:gcn}). 
Then, it employs a linear layer to project high-level hidden features for individual sentences to salience scores (Section~\ref{salience_estimation}).
Finally, it greedily produces relevant and non-redundant summaries by using sentence embeddings to detect similarities between candidate sentences and the current summary (Section~\ref{sum_process}).

The main contributions of this work are as follows:

\begin{itemize}
	\item We aggregate two types of sentences embeddings using a graph representation. They share different properties and are consequently complementary. The first one is trained on a large unrelated corpus to model general semantics among sentences, whereas the second is domain-specific to the dataset and learned during training. Together, they enable a model to be domain-independent as it can be applied easily on other domains. Moreover, it could be used for other tasks including detecting information cascades, query-focused summarization, keyphrase extraction and information retrieval.
	\item We devise a competitive multi-document summarization system, which does not need hand-crafted features nor additional annotated data. Moreover, the results are competitive for 665-byte and 100-word summaries. Usually, models are compared in one of the two settings but not both and thus lack comparability. 

\end{itemize}

\section{Method}
\label{sec:method}
Let $C$ denote a collection of related documents composed of a set of documents~$\{D_i|i \in [1,N]\}$ where $N$ is the number of documents. Moreover, each document~$D_i$ consists of a set of sentences $\{S_{i,j}|j \in [1,M]\}$, $M$ being the number of sentences in $D_i$. Given a collection of related documents~$C$, our goal is to produce a summary~$Sum$ using a subset of these in the input documents ordered in some way, such that $Sum = (S_{i_1,j_1},S_{i_2,j_2},...,S_{i_n,j_m})$. 

\begin{figure*}
  \includegraphics[width=1.0\linewidth]{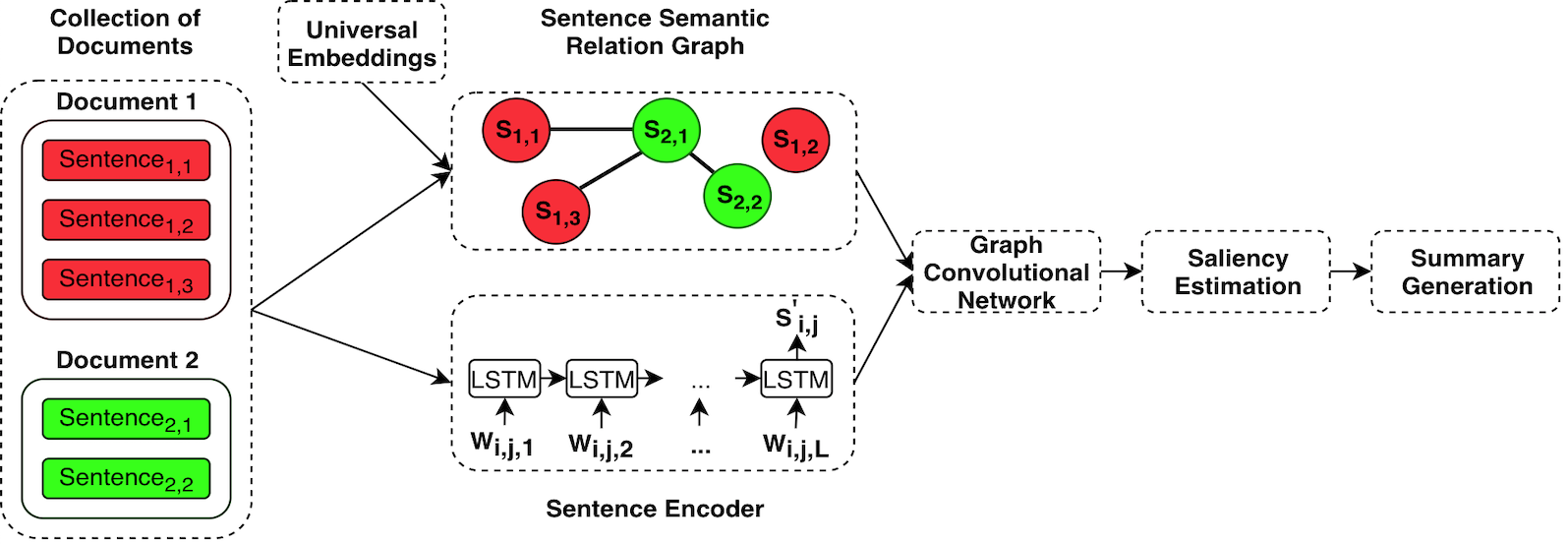}
  \caption{Overview of \textit{SemSentSum}. This illustration includes two documents in the collection, where the first one has three sentences and the second two. A sentence semantic relation graph is firstly built and each sentence node is processed by an encoder network at the same time. Thereafter, a single-layer graph convolutional network is applied on top and produces high-level hidden features for individual sentences. Then, salience scores are estimated using a linear layer and used to produce the final summary.}
  \label{fig:architecture}
\end{figure*}

In this section, we describe how \textit{SemSentSum} estimates the salience score of each sentence and how it selects a subset of these to create the final summary. The architecture of \textit{SemSentSum} is depicted in Figure~\ref{fig:architecture}.

In order to perform sentence selection, we first build our sentence semantic relation graph, where each vertex is a sentence and edges capture the semantic similarity among them. 
At the same time, each sentence is fed into a recurrent neural network, as a sentence encoder, to generate sentence embeddings using the last hidden states. 
A single-layer graph convolutional neural network is then applied on top, where the sentence semantic relation graph is the adjacency matrix and the sentence embeddings are the node features.
Afterward, a linear layer is used to project high-level hidden features for individual sentences to salience scores, representing how salient a sentence is with respect to the final summary. 
Finally, based on this, we devise an innovative greedy method that leverages sentence embeddings to detect redundant sentences and select sentences until reaching the summary length limit.

\subsection{Sentence Semantic Relation Graph}
\label{ssrg}
We model the semantic relationship among sentences using a graph representation. In this graph, each vertex is a sentence $S_{i,j}$ ($j$'th sentence of document $D_i$) from the collection documents $C$ and an undirected edge between $S_{i_u,j_u}$ and $S_{i_v,j_v}$ indicates their degree of similarity.
In order to compute the semantic similarity, we use the model of~\citet{pgj2017unsup} trained on the English Wikipedia corpus. In this manner, we incorporate general knowledge (i.e. not domain-specific) that will complete the specialized sentence embeddings obtained during training (see Section~\ref{sub:sentence_encoder}). 
We process sentences by their model and compute the cosine similarity between every sentence pair, resulting in a complete graph. 
However, having a complete graph alone does not allow the model to leverage the semantic structure across sentences significantly, as every sentence pair is connected, and likewise, a sparse graph does not contain enough information to exploit semantic similarities. 
Furthermore, all edges have a weight above zero, since it is very unlikely that two sentence embeddings are completely orthogonal.
To overcome this problem, we introduce an edge-removal-method, where ~every edge below a certain threshold $t_{sim}^g$ is removed in order to emphasize high sentence similarity. Nonetheless, $t_{sim}^g$ should not be too large, as we otherwise found the model to be prone to overfitting. 
After removing edges below $t_{sim}^g$, our sentence semantic relation graph is used as the adjacency matrix~$A$. The impact of $t_{sim}^g$ with different values is shown in Section~\ref{sec_res_and_disc}.

Based on our aforementioned hypothesis that a combination of general and genre-specific sentence embeddings is beneficial for the task of multi-document summarization, we further incorporate general sentence embeddings, pre-trained on Wikipedia entries, into edges between sentences. Additionally, we compute specialised sentence embeddings, which are related to the domains of the documents (see Section \ref{ablation_study}).

Note that 1) the pre-trained sentence embeddings are only used to compute the weights of the edges and are not used by the summarization model (as others are produced by the sentence encoder) and 2) the edge weights are static and do not change during training.

\subsection{Sentence Encoder}
\label{sub:sentence_encoder}

Given a list of documents $C$, we encode each document's sentence $S_{i,j}$, where each has at most $L$~words $(w_{i,j,1}, w_{i,j,2}, ..., w_{i,j,L})$. 
In our experiments, all words are kept and converted into word embeddings, which are then fed to the sentence encoder in order to compute specialized sentence embeddings $S'_{i,j}$.
We employ a single-layer forward recurrent neural network, using Long Short-Term Memory (LSTM) of~\cite{Hochreiter1997} as sentence encoder, where the sentence embeddings are extracted from the last hidden states. 
We then concatenate all sentence embeddings into a matrix~$X$ which constitutes the input node features that will be used by the graph convolutional network.

\subsection{Graph Convolutional Network}
\label{sub:gcn}

After having computed all sentence embeddings and the sentence semantic relation graph, we apply a single-layer Graph Convolutional Network (GCN) from~\citet{KipfW16}, in order to capture high-level hidden features for each sentence, encapsulating sentence information as well as the graph structure. 

We believe that our sentence semantic relation graph contains information not present in the data (via universal embeddings) and thus, we leverage this information by running a graph convolution on the first order neighborhood.

The GCN model takes as input the node features matrix~$X$ and a squared adjacency matrix~$A$. The former contains all sentence embeddings of the collection of documents, while the latter is our underlying sentence semantic relation graph.
It outputs hidden representations for each node that encode both local graph structure and nodes's features.
In order to take into account the sentences themselves during the information propagation, we add self-connections (i.e. the identity matrix) to $A$ such that $\tilde{A} = A + I$.
 
Subsequently, we obtain our sentence hidden features by using Equation~\ref{gcn_eq}.
\begin{equation}
\label{gcn_eq}
S''_{i,j} = \textrm{ELU}(\tilde{A}\textrm{ ELU}(\tilde{A}XW_0 + b_0)W_1 + b_1)	
\end{equation}
 where $W_i$ is the weight matrix of the $i$'th graph convolution layer and $b_i$ the bias vector. 
We choose the Exponential Linear Unit (ELU) activation function from \citet{ClevertUH15} due to its ability to handle the vanishing gradient problem, by pushing the mean unit activations close to zero and consequently facilitating the backpropagation. 
By using only one hidden layer, as we only have one input-to-hidden layer and one hidden-to-output layer, we limit the information propagation to the first order neighborhood.

\subsection{Saliency Estimation}
\label{salience_estimation}

We use a simple linear layer to estimate a salience score for each sentence and then normalize the scores via softmax and obtain our estimated salience score $S^s_{i,j}$.

\subsection{Training}

Our model \textit{SemSentSum} is trained in an end-to-end manner and minimizes the cross-entropy loss of Equation~\ref{crossentropy} between the salience score prediction and the ROUGE-1 $F_1$ score for each sentence.

\begin{equation}
\label{crossentropy}
	\mathcal{L} = -\sum_C \sum_{D \in C} \sum_{S \in D} F_1(S)\textrm{log}S^s
\end{equation}

$F_1(S)$ is computed as the ROUGE-1 $F_1$ score, unlike the common practice in the area of single and multi-document summarization as recall favors longer sentences whereas $F_1$ prevents this tendency. The scores are normalized via softmax.

\subsection{Summary Generation Process}
\label{sum_process}
While our model \textit{SemSentSum} provides estimated saliency scores, we use a greedy strategy to construct an informative and non-redundant summary~$Sum$. We first discard sentences having less than $9$ words, as in~\cite{erkan2004lexrank}, and then sort them in descending order of their estimated salience scores. We iteratively dequeue the sentence having the highest score and append it to the current summary $Sum$ if it is non-redundant with respect to the current content of $Sum$. We iterate until reaching the summary length limit.

To determine the similarity of a candidate sentence with the current summary, a sentence is considered as dissimilar if and only if the cosine similarity between its sentence embeddings and the embeddings of the current summary is below a certain threshold $t_{sim}^s$. 
We use the pre-trained model of~\citet{pgj2017unsup} to compute sentence as well as summary embeddings, similarly to the sentence semantic relation graph construction. 
Our approach is novel, since it focuses on the semantic sentence structures and captures similarity between sentence meanings, instead of focusing on word similarities only, like previous TF-IDF approaches (~\cite{hong2014improving,cao2015ranking,Yasunaga17,CaoLLW17}).

\section{Experiments}
\label{sec:experiments}

\subsection{Datasets}

We conduct experiments on the most commonly used datasets for multi-document summarization from the Document Understanding Conferences (DUC).\footnote{https://www-nlpir.nist.gov/projects/duc/guidelines.html} 
We use DUC 2001, 2002, 2003 and 2004 as the tasks of generic multi-document summarization, because they have been carried out during these years.
We use DUC 2001, 2002, 2003 and 2004 for generic multi-document summarization, where DUC~2001/2002 are used for training, DUC~2003 for validation and finally, DUC~2004 for testing, following the common practice.

\subsection{Evaluation Metric}
\label{sub:evaluation_metric}

For the evaluation, we use ROUGE \cite{Lin2004} with the official parameters of the DUC tasks and also truncate the summaries to 100 words for DUC 2001/2002/2003 and to 665 bytes for DUC 2004.\footnote{ROUGE-1.5.5 with options: -n 2 -m -u -c 95 -x -r 1000 -f A -p 0.5 -t 0 and -l 100 if using DUC 2001/2002/2003 otherwise -b 665.} Notably, we take ROUGE-1 and ROUGE-2 recall scores as the main metrics for comparison between produced summaries and golden ones as proposed by \cite{Owczarzak2012}. The goal of the ROUGE-N metric is to compute the ratio of the number of N-grams from the generated summary matching these of the human reference summaries.

\subsection{Model Settings}

To define the edge weights of our sentence semantic relation graph, we employ the~$600$-dimensional pre-trained unigram model of \citet{pgj2017unsup}, using English Wikipedia as source corpus. We keep only edges having a weight larger than~$t_{sim}^g = 0.5$ (tuned on the validation set). For word embeddings, the~$300$-dimensional pre-trained GloVe embeddings \cite{pennington2014glove} are used and fixed during training.
The output dimension of the sentence embeddings produced by the sentence encoder is the same as that of the word embeddings, i.e.~$300$. For the graph convolutional network, the number of hidden units is~$128$ and the size of the generated hidden feature vectors is also~$300$. We use a batch size of~$1$, a learning rate of~$0.0075$ using Adam optimizer~\cite{KingmaB14} with $\beta_1=0.9, \beta_2=0.999$ and $\epsilon=10^{-8}$. 
In order to make \textit{SemSentSum} generalize better, we use dropout \cite{Srivastava2014} of $0.2$, batch normalization \cite{ioffe2015batch}, clip the gradient norm at~$1.0$ if higher, add L2-norm regularizer with a regularization factor of~$10^{-12}$ and train using early stopping with a patience of~$10$ iterations. 
Finally, the similarity threshold $t_{sim}^s$ in the summary generation process is $0.8$ (tuned on the validation set). 

\subsection{Summarization Performance}

We train our model \textit{SemSentSum} on DUC 2001/2002, tune it on DUC 2003 and assess the performance on DUC 2004. In order to fairly compare \textit{SemSentSum} with other models available in the literature, experiments are conducted with summaries truncated to 665~bytes (official summary length in the DUC competition), but also with summaries with a length constraint of 100~words. 
To the best of our knowledge, we are the first to conduct experiments on both summary lengths and compare our model with other systems producing either 100 words or 665 bytes summaries.

\subsection{Sentence Semantic Relation Graph Construction}

We investigate different methods to build our sentence semantic relation graph and vary the value of $t_{sim}^g$ from $0.0$ to $0.75$ to study the impact of the threshold cut-off. 
Among these are:
\begin{enumerate}
	\item \textit{Cosine}: Using cosine similarity;
	\item \textit{Tf-idf}: Considering a node as the query and another as document. The weight corresponds to the cosine similarity between the query and the document;
	\item \textit{TextRank} \cite{Mihalcea04TextRank}: A weighted graph is created where nodes are sentences and edges defined by a similarity measure based on word overlap. Afterward, an algorithm similar to PageRank \cite{Pageetal98} is used to compute sentence importance and refined edge weights;
	\item \textit{LexRank} \cite{erkan2004lexrank}: An unsupervised multi-document summarizer based on the concept of eigenvector centrality in a graph of sentences to set up the edge weights;
	\item \textit{Approximate Discourse Graph} (ADG) \cite{christensen2013towards}: Approximation of a discourse graph where nodes are sentences and edges $(S_u,S_v)$ indicates sentence $S_v$ can be placed after $S_u$ in a coherent summary;
	\item \textit{Personalized ADG} (PADG) \cite{Yasunaga17}: Normalized version of ADG where sentence nodes are normalized over all edges.
\end{enumerate}

\subsection{Ablation Study}

In order to quantify the contribution of the different components of \textit{SemSentSum}, we try variations of our model by removing different modules one at a time. Our two main elements are the sentence encoder~(\textit{Sent}) and the graph convolutional neural network~(\textit{GCN}). When we omit \textit{Sent}, we substitute it with the pre-trained sentence embeddings used to build our sentence semantic relation graph.

\subsection{Results and Discussion}
\label{sec_res_and_disc}
Three dimensions are used to evaluate our model \textit{SemSentSum}: 1)~the summarization performance, to assess its capability 2)~the impact of the sentence semantic relation graph generation using various methods and different thresholds $t_{sim}^g$ 3)~an ablation study to analyze the importance of each component of \textit{SemSentSum}.

\paragraph{Summarization Performance}
We compare the results of \textit{SemSentSum} for both settings: 665~bytes and 100~words summaries. We only include models using the same parameters to compute the ROUGE-1/ROUGE-2 score and recall as metrics. 

\begin{table}
  \begin{tabular}{lcc}
    Model & ROUGE-1 & ROUGE-2  \\
    \hline
    MMR & $35.49$ & $7.50$ \\
    PV-DBOW+BS & $36.10$ & $6.77$ \\
    PG-MMR & $36.42$ & $9.36$ \\
    SVR & $36.18$ & $9.34$ \\
    G-Flow & $37.33$ & $8.74$ \\
    Peer 65 & $37.88$ & $9.18$\\
    R2N2 & $38.16$ & $9.52$ \\
    TCSum & $38.27$ & $\mathbf{9.66}$\\
    \hline
    SemSentSum & $\mathbf{39.12}$ & $9.59$ \\
  \end{tabular}
    \caption{Comparison of various models using ROUGE-1/ROUGE-2 on DUC 2004 with 665 bytes summaries.}
\label{sum_perf_665}
\end{table}

The results for 665~bytes summaries are reported in Table \ref{sum_perf_665}. We compare \textit{SemSentSum} with three types of model relying on either 1) sentence or document embeddings 2) various hand-crafted features or 3) additional data.
\begin{enumerate}
	\item For the first category, we significantly outperform \textbf{MMR}~\cite{bennani2018embedrank}, \textbf{PV-DBOW+BS}~\cite{mani2017multi} and \textbf{PG-MMR}~\cite{lebanoff2018adapting}. 
	Although their methods are based on embeddings to represent the meaning, it shows that using only various distance metrics or encoder-decoder architecture on these is not efficient for the task of multi-document summarization (as also shown in the Ablation Study). 
	We hypothesize that \textit{SemSentSum} performs better by leveraging pre-trained sentence embeddings and hence lowering the effects of data scarcity. 

	\item Systems based on hand-crafted features include a widely-used learning-based summarization method, built on support vector regression \textbf{SVR}~\cite{li2007multi}; a graph-based method based on approximating discourse graph \textbf{G-Flow}~\cite{christensen2013towards}; \textbf{Peer 65} which is the best peer systems participating in DUC evaluations; and the recursive neural network \textbf{R2N2} of \citet{cao2015ranking} that learns automatically combinations of hand-crafted features. 
	As can be seen, among these models completely dependent on hand-crafted features, \textit{SemSentSum} achieves highest performance on both ROUGE scores. 
	This denotes that using different linguistic and word-based features might not be enough to capture the semantic structures, in addition to being cumbersome to craft. 
	\item The last type of model is shown in \textbf{TCSum}~\cite{CaoLLW17} and uses transfer learning from a text classifier model, based on a domain-related dataset of $30\,000$ documents from New York Times (sharing the same topics of the DUC datasets). 
	In terms of ROUGE-1, \textit{SemSentSum} significantly outperforms \textbf{TCSum} and performs similarly on ROUGE-2 score. 
	This demonstrates that collecting more manually annotated data and training two models is unnecessary, in addition to being difficult to use in other domains, whereas \textit{SemSentSum} is fully data driven, domain-independent and usable in realistic scenarios.

\end{enumerate}

\begin{table}
  \begin{tabular}{lcc}
    Model & ROUGE-1 & ROUGE-2 \\
    \hline
    FreqSum & $35.30$ & $8.11$\\
    TsSum &  $35.88$ & $8.15$\\
    Cont. LexRank &  $35.95$ & $7.47$\\
	Centroid &  $36.41$ & $7.97$\\
	CLASSY04 &  $37.62$ & $8.96$\\
    CLASSY11 &  $37.22$ & $9.20$\\
    GreedyKL &  $37.98$ & $8.53$\\
    RegSum &  $38.57$ & $\mathbf{9.75}$\\
    GCN+PADG &  $38.23$ & $9.48$\\
    \hline
    SemSentSum & $\mathbf{38.72}$ & $9.69$ \\
  \end{tabular}
    \caption{Comparison of various models using ROUGE-1/2 on DUC 2004 with 100 words summaries.}
\label{sum_perf_100}
\end{table}

Table~\ref{sum_perf_100} depicts models producing 100~words summaries, all depending on hand-crafted features. We use as baselines \textbf{FreqSum}~\cite{Nenkova2006}; \textbf{TsSum}~\cite{conroy2006};  traditional graph-based approaches such as \textbf{Cont. LexRank}~\cite{erkan2004lexrank}; \textbf{Centroid}~\cite{radev2004centroid}; \textbf{CLASSY04}~\cite{conroy2004left}; its improved version \textbf{CLASSY11}~\cite{conroy2011classy} and the greedy model \textbf{GreedyKL}~\cite{Haghighi2009}. All of these models are significantly underperforming compared to \textit{SemSentSum}. In addition, we include state-of-the-art models: \textbf{RegSum}~\cite{hong2014improving} and \textbf{GCN+PADG}~\cite{Yasunaga17}. We outperform both in terms of ROUGE-1. For ROUGE-2 scores we achieve better results than \textbf{GCN+PADG} but without any use of domain-specific hand-crafted features and a much smaller and simpler model. Finally, \textbf{RegSum} achieves a similar ROUGE-2 score but computes sentence saliences based on word scores, incorporating a rich set of word-level and domain-specific features. Nonetheless, our model is competitive and does not depend on hand-crafted features due to its full data-driven nature and thus, it is not limited to a single domain.

Consequently, the experiments show that achieving good performance for multi-document summarization without hand-crafted features or additional data is clearly feasible and \textit{SemSentSum} produces competitive results without depending on these, is domain independent, fast to train and thus usable in real scenarios.

\paragraph{Sentence Semantic Relation Graph} Table \ref{ssrg_perf} shows the results of different methods to create the sentence semantic relation graph with various thresholds $t_{sim}^g$ for 665 bytes summaries (we obtain similar results for 100 words). A first observation is that using cosine similarity with sentence embeddings significantly outperforms all other methods for ROUGE-1 and ROUGE-2 scores, mainly because it relies on the semantic of sentences instead of their individual words. A second is that different methods evolve similarly: \textit{PADG, Textrank, Tf-idf} behave similarly to an U-shaped curve for both ROUGE scores while \textit{Cosine} is the only one having an inverted U-shaped curve. The reason for this behavior is a consequence of its distribution being similar to a normal distribution because it relies on the semantic instead of words, while the others are more skewed towards zero. This confirms our hypothesis that 1)~having a complete graph does not allow the model to leverage much the semantic 2)~a sparse graph might not contain enough information to exploit similarities. Finally, \textit{Lexrank} and \textit{ADG} have different trends between both ROUGE scores.

\begin{table*}
  \begin{tabular}{lccccccccccc}
    & & \multicolumn{4}{c}{ROUGE-1} & & & \multicolumn{4}{c}{ROUGE-2}  \\
    Method & $t_{sim}^g$ & $0.0$ & $0.25$ & $0.5$ & $0.75$ & & $t_{sim}^g$ & $0.0$ & $0.25$ & $0.5$ & $0.75$\\
    \hline
    Cosine & & $38.49{*}$ & $38.61{*}$ & $\mathbf{39.12}\textrm{ }$ & $35.54{*}$ & & & $9.11{*}$ & $9.07{*}$ & $\mathbf{9.59}\textrm{ }$ & $7.12{*}$\\
    Tf-idf & & $36.80{*}$ & $36.23{*}$ & $35.26{*}$ & $35.71{*}$ & & & $7.84{*}$ & $7.78{*}$ & $7.07{*}$ & $7.46{*}$\\
    Textrank & & $35.66{*}$ & $34.75{*}$ & $35.41{*}$ & $35.69{*}$ & & & $7.83{*}$ & $7.17{*}$ & $7.20{*}$ & $7.54{*}$\\
    Lexrank & & $37.04{*}$ & $36.43{*}$ & $36.27{*}$ & $35.65{*}$ & & & $7.90{*}$ & $8.01{*}$ & $7.64{*}$ & $7.61{*}$\\
    ADG & & $35.48{*}$ & $34.79{*}$ & $34.78{*}$ & $35.40{*}$ & & & $6.96{*}$ & $7.03{*}$ & $7.01{*}$ & $7.32{*}$\\
    PADG & & $36.81{*}$ & $36.23{*}$ & $35.26{*}$ & $35.71{*}$ & & & $7.84{*}$ & $7.78{*}$ & $7.07{*}$ & $7.46{*}$\\
  \end{tabular}
    \caption{ROUGE-1/2 for various methods to build the sentence semantic relation graph. A score significantly different (according to a Welch Two Sample t-test, $p = 0.001$) than cosine similarity ($t_{sim}^g=0.5$) is denoted by ${*}$.}
  \label{ssrg_perf}
\end{table*}

\paragraph{Ablation Study}
\label{ablation_study}
We quantify the contribution of each module of \textit{SemSentSum} in Table \ref{abl_study} for 665~bytes summaries (we obtain similar results for 100 words). 
Removing the sentence encoder produces slightly lower results. 
This shows that the sentence semantic relation graph captures semantic attributes well, while the fine-tuned sentence embeddings obtained via the encoder help boost the performance, making these methods complementary. 
By disabling only the graph convolutional layer, a drastic drop in terms of performance is observed, which emphasizes that the relationship among sentences is indeed important and not present in the data itself. 
Therefore, our sentence semantic relation graph is able to capture sentence similarities by analyzing the semantic structures. 
Interestingly, if we remove the sentence encoder in addition to the graph convolutional layer, similar results are achieved. This confirms that alone, the sentence encoder is not able to compute an efficient representation of sentences for the task of multi-document summarization, probably due to the poor size of the DUC datasets. 
Finally, we can observe that the use of sentence embeddings only results in similar performance to the baselines, which rely on sentence or document embeddings~\cite{bennani2018embedrank,mani2017multi}.

\begin{table}
  \begin{tabular}{lcc}
    Model & ROUGE-1 & ROUGE-2  \\
    \hline
    \textit{SemSentSum} & $\mathbf{39.12}\textrm{ }$ & $\mathbf{9.59}\textrm{ }$\\
    - w/o Sent & $38.38{*}$ & $9.11{*}$\\
    - w/o GCN & $35.88{*}$ & $7.33{*}$\\
    - w/o GCN,Sent & $35.89{*}$ & $7.24{*}$\\
  \end{tabular}
  \caption{Ablation test. \textit{Sent} is the sentence encoder and \textit{GCN} the graph convolutional network. According to a Welch Two Sample t-test ($p = 0.001$), a score significantly different than \textit{SemSentSum} is denoted by ${*}$.}
    \label{abl_study}
\end{table}

\section{Related Work}
\label{sec:related_work}

The idea of using multiple embeddings has been employed at the word level. 
\citet{kiela2018context} use an attention mechanism to combine the embeddings for each word for the task of natural language inference. \citet{xu2018double, bollegala2015unsupervised} concatenate the embeddings of each word into a vector before feeding a neural network for the tasks of aspect extraction and sentiment analysis. 
To our knowledge, we are the first to combine multiple types of sentence embeddings.
 
Extractive multi-document summarization has been addressed by a large range of approaches. 
Several of them employ graph-based methods. 
\citet{Radev2000} introduced a cross-document structure theory, as a basis for multi-document summarization. 
\citet{erkan2004lexrank} proposed LexRank, an unsupervised multi-document summarizer based on the concept of eigenvector centrality in a graph of sentences. 
Other works exploit shallow or deep features from the graph's topology \cite{Wan2006,ANTIQUEIRA2009584}. 
\citet{Wan2008} pairs graph-based methods (e.g. random walk) with clustering. 
\citet{Mei20101} improved results by using a reinforced random walk model to rank sentences and keep non-redundant ones. 
The system by \citet{christensen2013towards} does sentence selection, while balancing coherence and salience and by building a graph that approximates discourse relations across sentences \cite{mann88b}.

Besides graph-based methods, other viable approaches include Maximum Marginal Relevance \cite{Carbonell1998}, which uses a greedy approach to select sentences and considers the tradeoff between relevance and redundancy; support vector regression \cite{li2007multi}; conditional random field \cite{Galley2006}; or hidden markov model \cite{conroy2004left}. 
Yet other approaches rely on n-grams regression as in \citet{li2013using}. 
More recently, \citet{cao2015ranking} built a recursive neural network, which tries to automatically detect combination of hand-crafted features. 
\citet{CaoLLW17} employ a neural model for text classification on a large manually annotated dataset and apply transfer learning for multi-document summarization afterward.

The work most closely related to ours is \cite{Yasunaga17}. 
They create a normalized version of the approximate discourse graph \cite{christensen2013towards}, based on hand-crafted features, where sentence nodes are normalized over all the incoming edges. 
They then employ a deep neural network, composed of a sentence encoder, three graph convolutional layers, one document encoder and an attention mechanism. 
Afterward, they greedily select sentences using TF-IDF similarity to detect redundant sentences. 
Our model differs in four ways: 1) we build our sentence semantic relation graph by using pre-trained sentence embeddings with cosine similarity, where neither heavy preprocessing, nor hand-crafted features are necessary. Thus, our model is fully data-driven and domain-independent unlike other systems. In addition, the sentence semantic relation graph could be used for other tasks than multi-document summarization, such as detecting information cascades, query-focused summarization, keyphrase extraction or information retrieval, as it is not composed of hand-crafted features. 2) \textit{SemSentSum} is much smaller and consequently has fewer parameters as it only uses a sentence encoder and a single convolutional layer. 3) The loss function is based on ROUGE-1 $F_1$ score instead of recall to prevent the tendency of choosing longer sentences. 4) Our method for summary generation is also different and novel as we leverage sentence embeddings to compute the similarity between a candidate sentence and the current summary instead of TF-IDF based approaches.

\section{Conclusion}
\label{sec:conclusion}

In this work, we propose a method to combine two types of sentence embeddings: 1) universal embeddings, pre-trained on a large corpus such as Wikipedia and incorporating general semantic structures across sentences and 2) domain-specific embeddings, learned during training. We merge them together by using a graph convolutional network that eliminates the need of hand-crafted features or additional annotated data.

We introduce a fully data-driven model \textit{SemSentSum} that achieves competitive results for multi-document summarization on both kind of summary length (665~bytes and 100~words summaries), without requiring hand-crafted features or additional annotated data.

As \textit{SemSentSum} is domain-independent, we believe that our sentence semantic relation graph and model can be used for other tasks including detecting information cascades, query-focused summarization, keyphrase extraction and information retrieval. 
In addition, we plan to leave the weights of the sentence semantic relation graph dynamic during training, and to integrate an attention mechanism directly into the graph.

\section*{Acknowledgments}

We thank Michaela Benk for proofreading and helpful advice.

\bibliographystyle{acl_natbib}

\begin{thebibliography}{40}
\expandafter\ifx\csname natexlab\endcsname\relax\def\natexlab#1{#1}\fi

\bibitem[{Antiqueira et~al.(2009)Antiqueira, Oliveira, da~Fontoura~Costa, and
  das Graças Volpe~Nunes}]{ANTIQUEIRA2009584}
Lucas Antiqueira, Osvaldo~N. Oliveira, Luciano da~Fontoura~Costa, and Maria das
  Graças Volpe~Nunes. 2009.
\newblock \href {https://doi.org/https://doi.org/10.1016/j.ins.2008.10.032} {A
  complex network approach to text summarization}.
\newblock \emph{Information Sciences}, 179(5):584 -- 599.
\newblock Special Section - Quantum Structures: Theory and Applications.

\bibitem[{Bennani-Smires et~al.(2018)Bennani-Smires, Musat, Hossmann,
  Baeriswyl, and Jaggi}]{bennani2018embedrank}
Kamil Bennani-Smires, Claudiu-Cristian Musat, Andreea Hossmann, Michael
  Baeriswyl, and Martin Jaggi. 2018.
\newblock Simple unsupervised keyphrase extraction using sentence embeddings.
\newblock \emph{Proceedings of the 22nd Conference on Computational Natural
  Language Learning, CoNLL 2018, Brussels, Belgium, October 31 - November 1,
  2018}, pages 221--229.

\bibitem[{Bollegala et~al.(2015)Bollegala, Maehara, and ichi
  Kawarabayashi}]{bollegala2015unsupervised}
Danushka Bollegala, Takanori Maehara, and Ken ichi Kawarabayashi. 2015.
\newblock Unsupervised cross-domain word representation learning.
\newblock \emph{Proc. of the Annual Conference of the Association for
  Computational Linguistics (ACL) and International Joint Conferences on
  Natural Language Processing (IJCNLP)}.

\bibitem[{Cao et~al.(2017)Cao, Li, Li, and Wei}]{CaoLLW17}
Ziqiang Cao, Wenjie Li, Sujian Li, and Furu Wei. 2017.
\newblock \href {http://aaai.org/ocs/index.php/AAAI/AAAI17/paper/view/14525}
  {Improving multi-document summarization via text classification}.
\newblock In \emph{Proceedings of the Thirty-First {AAAI} Conference on
  Artificial Intelligence, February 4-9, 2017, San Francisco, California,
  {USA.}}, pages 3053--3059.

\bibitem[{Cao et~al.(2015)Cao, Wei, Dong, Li, and Zhou}]{cao2015ranking}
Ziqiang Cao, Furu Wei, Li~Dong, Sujian Li, and Ming Zhou. 2015.
\newblock \href {http://dl.acm.org/citation.cfm?id=2886521.2886620} {Ranking
  with recursive neural networks and its application to multi-document
  summarization}.
\newblock In \emph{Proceedings of the Twenty-Ninth AAAI Conference on
  Artificial Intelligence}, AAAI'15, pages 2153--2159. AAAI Press.

\bibitem[{Carbonell and Goldstein(1998)}]{Carbonell1998}
Jaime Carbonell and Jade Goldstein. 1998.
\newblock \href {https://doi.org/10.1145/290941.291025} {The use of mmr,
  diversity-based reranking for reordering documents and producing summaries}.
\newblock In \emph{Proceedings of the 21st Annual International ACM SIGIR
  Conference on Research and Development in Information Retrieval}, SIGIR '98,
  pages 335--336, New York, NY, USA. ACM.

\bibitem[{Christensen et~al.(2013)Christensen, Soderland, Etzioni
  et~al.}]{christensen2013towards}
Janara Christensen, Stephen Soderland, Oren Etzioni, et~al. 2013.
\newblock Towards coherent multi-document summarization.
\newblock In \emph{Proceedings of the 2013 conference of the North American
  chapter of the association for computational linguistics: Human language
  technologies}, pages 1163--1173.

\bibitem[{Clevert et~al.(2016)Clevert, Unterthiner, and
  Hochreiter}]{ClevertUH15}
Djork{-}Arn{\'{e}} Clevert, Thomas Unterthiner, and Sepp Hochreiter. 2016.
\newblock \href {http://arxiv.org/abs/1511.07289} {Fast and accurate deep
  network learning by exponential linear units (elus)}.
\newblock \emph{International Conference on Learning Representations}.

\bibitem[{Conroy et~al.(2004)Conroy, Schlesinger, Goldstein, and
  O’leary}]{conroy2004left}
John~M Conroy, Judith~D Schlesinger, Jade Goldstein, and Dianne~P O’leary.
  2004.
\newblock Left-brain/right-brain multi-document summarization.
\newblock In \emph{Proceedings of the Document Understanding Conference (DUC
  2004)}.

\bibitem[{Conroy et~al.(2011)Conroy, Schlesinger, Kubina, Rankel, and
  O'Leary}]{conroy2011classy}
John~M Conroy, Judith~D Schlesinger, Jeff Kubina, Peter~A Rankel, and Dianne~P
  O'Leary. 2011.
\newblock Classy 2011 at tac: Guided and multi-lingual summaries and evaluation
  metrics.
\newblock \emph{TAC}, 11:1--8.

\bibitem[{Conroy et~al.(2006)Conroy, Schlesinger, and O'Leary}]{conroy2006}
John~M. Conroy, Judith~D. Schlesinger, and Dianne~P. O'Leary. 2006.
\newblock \href {http://dl.acm.org/citation.cfm?id=1273073.1273093}
  {Topic-focused multi-document summarization using an approximate oracle
  score}.
\newblock In \emph{Proceedings of the COLING/ACL on Main Conference Poster
  Sessions}, COLING-ACL '06, pages 152--159, Stroudsburg, PA, USA. Association
  for Computational Linguistics.

\bibitem[{Erkan and Radev(2004)}]{erkan2004lexrank}
G{\"u}nes Erkan and Dragomir~R Radev. 2004.
\newblock Lexrank: Graph-based lexical centrality as salience in text
  summarization.
\newblock \emph{Journal of Artificial Intelligence Research}, 22:457--479.

\bibitem[{Galley(2006)}]{Galley2006}
Michel Galley. 2006.
\newblock \href {http://dl.acm.org/citation.cfm?id=1610075.1610126} {A
  skip-chain conditional random field for ranking meeting utterances by
  importance}.
\newblock In \emph{Proceedings of the 2006 Conference on Empirical Methods in
  Natural Language Processing}, EMNLP '06, pages 364--372, Stroudsburg, PA,
  USA. Association for Computational Linguistics.

\bibitem[{Haghighi and Vanderwende(2009)}]{Haghighi2009}
Aria Haghighi and Lucy Vanderwende. 2009.
\newblock \href {http://dl.acm.org/citation.cfm?id=1620754.1620807} {Exploring
  content models for multi-document summarization}.
\newblock In \emph{Proceedings of Human Language Technologies: The 2009 Annual
  Conference of the North American Chapter of the Association for Computational
  Linguistics}, NAACL '09, pages 362--370, Stroudsburg, PA, USA. Association
  for Computational Linguistics.

\bibitem[{Hochreiter and Schmidhuber(1997)}]{Hochreiter1997}
Sepp Hochreiter and J\"{u}rgen Schmidhuber. 1997.
\newblock \href {https://doi.org/10.1162/neco.1997.9.8.1735} {Long short-term
  memory}.
\newblock \emph{Neural Comput.}, 9(8):1735--1780.

\bibitem[{Hong and Nenkova(2014)}]{hong2014improving}
Kai Hong and Ani Nenkova. 2014.
\newblock Improving the estimation of word importance for news multi-document
  summarization.
\newblock pages 712--721.

\bibitem[{Ioffe and Szegedy(2015)}]{ioffe2015batch}
Sergey Ioffe and Christian Szegedy. 2015.
\newblock Batch normalization: Accelerating deep network training by reducing
  internal covariate shift.
\newblock In \emph{International Conference on Machine Learning}, pages
  448--456.

\bibitem[{Kiela et~al.(2018)Kiela, Wang, and Cho}]{kiela2018context}
Douwe Kiela, Changhan Wang, and Kyunghyun Cho. 2018.
\newblock Dynamic meta-embeddings for improved sentence representations.
\newblock \emph{Proceedings of the 2018 Conference on Empirical Methods in
  Natural Language Processing (EMNLP)}.

\bibitem[{Kingma and Ba(2015)}]{KingmaB14}
Diederik~P. Kingma and Jimmy Ba. 2015.
\newblock \href
  {http://dblp.uni-trier.de/db/journals/corr/corr1412.html#KingmaB14} {Adam: A
  method for stochastic optimization.}
\newblock \emph{International Conference on Learning Representations}.

\bibitem[{Kipf and Welling(2017)}]{KipfW16}
Thomas~N. Kipf and Max Welling. 2017.
\newblock \href {http://arxiv.org/abs/1609.02907} {Semi-supervised
  classification with graph convolutional networks}.
\newblock \emph{International Conference on Learning Representations}.

\bibitem[{Lebanoff et~al.(2018)Lebanoff, Song, and Liu}]{lebanoff2018adapting}
Logan Lebanoff, Kaiqiang Song, and Fei Liu. 2018.
\newblock Adapting the neural encoder-decoder framework from single to
  multi-document summarization.
\newblock \emph{Proceedings of the 2018 Conference on Empirical Methods in
  Natural Language Processing}.

\bibitem[{Li et~al.(2013)Li, Qian, and Liu}]{li2013using}
Chen Li, Xian Qian, and Yang Liu. 2013.
\newblock Using supervised bigram-based ilp for extractive summarization.
\newblock In \emph{Proceedings of the 51st Annual Meeting of the Association
  for Computational Linguistics (Volume 1: Long Papers)}, volume~1, pages
  1004--1013.

\bibitem[{Li et~al.(2007)Li, Ouyang, Wang, and Sun}]{li2007multi}
Sujian Li, You Ouyang, Wei Wang, and Bin Sun. 2007.
\newblock Multi-document summarization using support vector regression.
\newblock In \emph{In Proceedings of DUC.} Citeseer.

\bibitem[{Lin(2004)}]{Lin2004}
C.~Y. Lin. 2004.
\newblock {ROUGE}: A package for automatic evaluation of summaries.
\newblock In \emph{Proceedings of the Workshop on Text Summarization Branches
  Out (WAS)}, Barcelona, Spain.

\bibitem[{Mani et~al.(2017)Mani, Verma, and Dey}]{mani2017multi}
Kaustubh Mani, Ishan Verma, and Lipika Dey. 2017.
\newblock Multi-document summarization using distributed bag-of-words model.
\newblock \emph{arXiv preprint arXiv:1710.02745}.

\bibitem[{Mann and Thompson(1988)}]{mann88b}
William~C. Mann and Sandra~A. Thompson. 1988.
\newblock Rhetorical structure theory: Toward a functional theory of text
  organization.
\newblock \emph{Text}, 8(3):243--281.

\bibitem[{Mei et~al.(2010)Mei, Guo, and Radev}]{Mei20101}
Qiaozhu Mei, Jian Guo, and Dragomir Radev. 2010.
\newblock \href {https://doi.org/10.1145/1835804.1835931} {Divrank: The
  interplay of prestige and diversity in information networks}.
\newblock In \emph{Proceedings of the 16th ACM SIGKDD International Conference
  on Knowledge Discovery and Data Mining}, KDD '10, pages 1009--1018, New York,
  NY, USA. ACM.

\bibitem[{Mihalcea and Tarau(2004)}]{Mihalcea04TextRank}
R.~Mihalcea and P.~Tarau. 2004.
\newblock {TextRank}: Bringing order into texts.
\newblock In \emph{Proceedings of {EMNLP-04}and the 2004 Conference on
  Empirical Methods in Natural Language Processing}.

\bibitem[{Nenkova et~al.(2006)Nenkova, Vanderwende, and McKeown}]{Nenkova2006}
Ani Nenkova, Lucy Vanderwende, and Kathleen McKeown. 2006.
\newblock \href {https://doi.org/10.1145/1148170.1148269} {A compositional
  context sensitive multi-document summarizer: Exploring the factors that
  influence summarization}.
\newblock In \emph{Proceedings of the 29th Annual International ACM SIGIR
  Conference on Research and Development in Information Retrieval}, SIGIR '06,
  pages 573--580, New York, NY, USA. ACM.

\bibitem[{Owczarzak et~al.(2012)Owczarzak, Conroy, Dang, and
  Nenkova}]{Owczarzak2012}
Karolina Owczarzak, John~M. Conroy, Hoa~Trang Dang, and Ani Nenkova. 2012.
\newblock \href {http://dl.acm.org/citation.cfm?id=2391258.2391259} {An
  assessment of the accuracy of automatic evaluation in summarization}.
\newblock In \emph{Proceedings of Workshop on Evaluation Metrics and System
  Comparison for Automatic Summarization}, pages 1--9, Stroudsburg, PA, USA.
  Association for Computational Linguistics.

\bibitem[{Page et~al.(1998)Page, Brin, Motwani, and Winograd}]{Pageetal98}
L.~Page, S.~Brin, R.~Motwani, and T.~Winograd. 1998.
\newblock \href {citeseer.nj.nec.com/page98pagerank.html} {The pagerank
  citation ranking: Bringing order to the web}.
\newblock In \emph{Proceedings of the 7th International World Wide Web
  Conference}, pages 161--172, Brisbane, Australia.

\bibitem[{Pagliardini et~al.(2018)Pagliardini, Gupta, and Jaggi}]{pgj2017unsup}
Matteo Pagliardini, Prakhar Gupta, and Martin Jaggi. 2018.
\newblock {Unsupervised Learning of Sentence Embeddings using Compositional
  n-Gram Features}.
\newblock In \emph{NAACL 2018 - Conference of the North American Chapter of the
  Association for Computational Linguistics}.

\bibitem[{Pennington et~al.(2014)Pennington, Socher, and
  Manning}]{pennington2014glove}
Jeffrey Pennington, Richard Socher, and Christopher~D. Manning. 2014.
\newblock \href {http://www.aclweb.org/anthology/D14-1162} {Glove: Global
  vectors for word representation}.
\newblock In \emph{Empirical Methods in Natural Language Processing (EMNLP)},
  pages 1532--1543.

\bibitem[{Radev(2000)}]{Radev2000}
Dragomir~R. Radev. 2000.
\newblock \href {https://doi.org/10.3115/1117736.1117745} {A common theory of
  information fusion from multiple text sources step one: Cross-document
  structure}.
\newblock In \emph{Proceedings of the 1st SIGdial Workshop on Discourse and
  Dialogue - Volume 10}, SIGDIAL '00, pages 74--83, Stroudsburg, PA, USA.
  Association for Computational Linguistics.

\bibitem[{Radev et~al.(2004)Radev, Jing, Sty{\'s}, and Tam}]{radev2004centroid}
Dragomir~R Radev, Hongyan Jing, Ma{\l}gorzata Sty{\'s}, and Daniel Tam. 2004.
\newblock Centroid-based summarization of multiple documents.
\newblock \emph{Information Processing \& Management}, 40(6):919--938.

\bibitem[{Srivastava et~al.(2014)Srivastava, Hinton, Krizhevsky, Sutskever, and
  Salakhutdinov}]{Srivastava2014}
Nitish Srivastava, Geoffrey Hinton, Alex Krizhevsky, Ilya Sutskever, and Ruslan
  Salakhutdinov. 2014.
\newblock \href {http://dl.acm.org/citation.cfm?id=2627435.2670313} {Dropout: A
  simple way to prevent neural networks from overfitting}.
\newblock \emph{J. Mach. Learn. Res.}, 15(1):1929--1958.

\bibitem[{Wan and Yang(2006)}]{Wan2006}
Xiaojun Wan and Jianwu Yang. 2006.
\newblock \href {http://dl.acm.org/citation.cfm?id=1614049.1614095} {Improved
  affinity graph based multi-document summarization}.
\newblock In \emph{Proceedings of the Human Language Technology Conference of
  the NAACL, Companion Volume: Short Papers}, NAACL-Short '06, pages 181--184,
  Stroudsburg, PA, USA. Association for Computational Linguistics.

\bibitem[{Wan and Yang(2008)}]{Wan2008}
Xiaojun Wan and Jianwu Yang. 2008.
\newblock \href {https://doi.org/10.1145/1390334.1390386} {Multi-document
  summarization using cluster-based link analysis}.
\newblock In \emph{Proceedings of the 31st Annual International ACM SIGIR
  Conference on Research and Development in Information Retrieval}, SIGIR '08,
  pages 299--306, New York, NY, USA. ACM.

\bibitem[{Xu et~al.(2018)Xu, Liu, Shu, and Yu}]{xu2018double}
Hu~Xu, Bing Liu, Lei Shu, and Philip~S Yu. 2018.
\newblock Double embeddings and cnn-based sequence labeling for aspect
  extraction.
\newblock \emph{Proceedings of the 56th Annual Meeting of the Association for
  Computational Linguistics}.

\bibitem[{Yasunaga et~al.(2017)Yasunaga, Zhang, Meelu, Pareek, Srinivasan, and
  Radev}]{Yasunaga17}
Michihiro Yasunaga, Rui Zhang, Kshitijh Meelu, Ayush Pareek, Krishnan
  Srinivasan, and Dragomir~R. Radev. 2017.
\newblock Graph-based neural multi-document summarization.
\newblock In \emph{Proceedings of CoNLL-2017}. Association for Computational
  Linguistics.

\end{thebibliography}

\end{document}